\documentclass[letterpaper, 10 pt, conference]{ieeeconf}  % Comment this line out if you need a4paper

\IEEEoverridecommandlockouts                              % This command is only needed if 
\usepackage[colorlinks=true,linkcolor=blue,citecolor=blue,urlcolor=blue]{hyperref}
\usepackage{graphics} % for pdf, bitmapped graphics files
\usepackage{epsfig} % for postscript graphics files
\usepackage{times} % assumes new font selection scheme installed
\usepackage{amsmath} % assumes amsmath package installed
\usepackage{amssymb}  % assumes amsmath package installed
\usepackage{booktabs}
\usepackage{multirow}
\usepackage{array}
\usepackage{caption}
\usepackage{subcaption}
\usepackage{graphicx} 
\usepackage{cleveref}
\usepackage{xcolor}
\usepackage{algorithm}
\usepackage{algpseudocode}
\title{\LARGE \bf
Coordinated Motion Planning for Multi-Arm Systems via \\ Iterative LQ Games
}

\author{Junyoung Kim, Hanwen Ren, Lei Zhang, and Ahmed H. Qureshi %
}

\begin{document}

% \maketitle
\thispagestyle{empty}
\pagestyle{empty}

%%%%%%%%%%%%%%%%%%%%%%%%%%%%%%%%%%%%%%%%%%%%%%%%%%%%%%%%%%%%%%%%%%%%%%%%%%%%%%%%

\twocolumn[{
\begin{@twocolumnfalse}
\maketitle
\begin{center}
    \captionsetup{type=figure}
    \includegraphics[trim={0.65cm 0.2cm 0.7cm 0.2cm},clip,width=17cm]{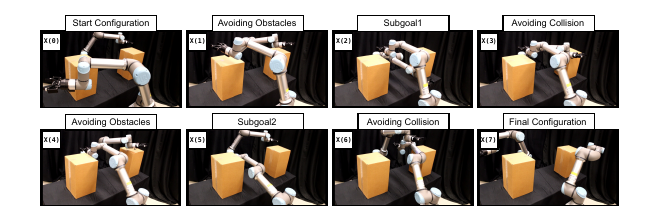}
    \captionof{figure}{The figure shows a multi-agent motion planning with static obstacles. Each key-frame shows the description of each movement with the time step on the right corner. Our ILQ-Arm finds collision free solution in 1.065 seconds, while the execution lasted 110 seconds.}
    \label{fig:real_exp1}%
\end{center}
\end{@twocolumnfalse}
}]
\begingroup
\renewcommand{\thefootnote}{}
\footnotetext{Junyoung Kim, Hanwen Ren, Lei Zhang and Ahmed H. Qureshi are with the Department of Computer Science, Purdue University, West Lafayette, IN, USA, 47907. Email {\tt\small$\{$kim3722, ren221, zhan5814, ahqureshi$\}@$purdue.edu}.}
\endgroup

\begin{abstract}

Multi-agent motion planning for high-degree-of-freedom robotics manipulators in shared workspaces remains a fundamental yet challenging problem. Centralized planners often suffer from poor scalability, while decentralized approaches face robustness and safety concerns. Game-theoretic formulations offer a promising approach for modeling agent interactions, potentially overcoming these limitations. However, their application to articulated multi-arm systems remains limited. This paper presents an iterative Linear Quadratic (LQ) game framework for multi-manipulator motion planning, where each manipulator is modeled as an independent agent optimizing its own objective while interacting with other agents based on shared global states and collision constraints. The method solves a series of local LQ games by linearizing the dynamics and approximating the cost around a nominal trajectory, with Riccati backward recursions yielding feedback Nash strategies. To address the challenges of articulated systems, we incorporate differentiable penalties for self-collision and inter-arm collision into the optimization pipeline, enabling coordinated, collision-aware trajectory generation. Experiments demonstrate that our framework produces smooth, safe, and efficient trajectories in high-dimensional settings, outperforming traditional methods. This highlights the effectiveness of differential game formulations for multi-robot manipulation.

\end{abstract}

%%%%%%%%%%%%%%%%%%%%%%%%%%%%%%%%%%%%%%%%%%%%%%%%%%%%%%%%%%%%%%%%%%%%%%%%%%%%%%%%
\section{INTRODUCTION}
Multi-Agent Motion Planning (MAMP) for multiple robotic manipulators is a fundamental challenge in various underlying robotics applications, including collaborative assembly~\cite{dogar2012planning, kruger2009cooperation}, warehouse automation~\cite{wurman2008coordinating, ren2026multi}, surgical robotics~\cite{kazanzides2014open}, and human-robot interaction~\cite{mainprice2013human, keshari2022cograsp}. These applications require the coordination of multiple high-degree-of-freedom (DoF) arms within shared workspaces, where the robots must not only achieve their respective tasks but also ensure safe operation by avoiding collisions. This makes MAMP crucial for the effective deployment of multi-robot systems in dynamic environments.

Existing motion planning algorithms can be categorized into centralized and decentralized approaches. Classical centralized planners rely on a central controller to generate actions for all agents, optimizing over the joint state space to provide globally consistent solutions. While these methods work well in low-dimensional settings, they struggle with scalability and real-time performance in multi-manipulator scenarios. In contrast, decentralized planners allow each agent to plan independently, improving computational efficiency. However, they often face challenges related to coordination, robustness, and safety, as they rely on policy assumptions or predictions regarding the actions of other agents.

To address these challenges, game-theoretic formulations provide a promising framework for modeling interactions among agents. In differential games, each agent minimizes its own cost function while considering the interactions with other agents based on the shared global state. A Nash equilibrium is achieved when each agent's strategy is the best response to the strategies of the others, meaning no agent can improve their payoff by unilaterally changing their strategy. This approach has been widely applied in domains such as autonomous driving~\cite{fisac2019hierarchical}, pursuit-evasion scenarios~\cite{isaacs1999differential, mitchell2005time}, and multi-robotic systems~\cite{fridovich2020iterative}. Among the game-theoretic solution methods, Linear Quadratic (LQ) games~\cite{bacsar1998dynamic, engwerda2005lq} offer closed-form solutions for the Nash equilibria under linear dynamics and quadratic costs. To extend to nonlinear systems, iterative Linear-Quadratic (ILQ) games~\cite{fridovich2020efficient} linearize the dynamics and approximate the cost around a nominal trajectory. This results in a series of local LQ games, which can be efficiently solved via Riccati backward recursions, yielding time-varying state feedback Nash strategies. ILQ game formulations have demonstrated strong performance in multi-vehicle and interactive control scenarios due to their computational efficiency and compatibility with feedback policies~\cite{fisac2019hierarchical}.

Despite the advantages of ILQ methods in low-dimensional systems, such as ground vehicles and bicycle kinematics models, their application to high-DOF robotic manipulators remains unexplored. Multi-arm manipulation introduces additional challenges, including complex articulated kinematics, self-collision constraints, and tightly coupled geometric interaction. These challenges are difficult to model within traditional ILQ frameworks, which struggle to handle the complexity of collision penalties and the numerical conditioning required for second-order optimization procedures. Extending the ILQ framework to effectively model and solve multi-agent motion planning (MAMP) for robotic manipulators is a critical practical challenge that remains largely underexplored from a game-theoretic perspective.

To address these challenges, we present ILQ-Arm, a motion planning framework for multi-manipulator systems based on the ILQ game framework. In our approach, each manipulator is modeled as an independent agent seeking to minimize its own objective while interacting with other agents based on shared global states and collision constraints. Thus, our ILQ-Arm is a centralized solver that computes decentralized, per-agent feedback strategies by jointly solving each agent's local game around the shared global state. By leveraging the computational efficiency of ILQ games, our method efficiently computes time-varying state-feedback Nash strategies for high-DoF articulated systems. Unlike prior ILQ game formulations, which focus on low-dimensional agents with simplified geometries, ILQ-Arm explicitly incorporates differentiable self-collision and inter-arm collision penalties tailored to complex manipulator kinematics. This enables coordinated, collision-aware motion generation within a unified second-order optimization framework. Our key contributions are summarized as follows.

\begin{itemize}
    \item A differential game formulation for multi-manipulator motion planning, where each arm is modeled as an optimization agent, and feedback Nash strategies are computed using an ILQ game procedure.
    \item An integration of differentiable geometric collision penalties, including self-collision modeling, within the ILQ game backward–forward optimization pipeline.
    \item A systematic analysis of cost design, including quadratic control regularization and goal costs, demonstrating their impact on convergence, safety, and trajectory quality.
    \item Extensive validation on challenging multi-arm planning scenarios, involving varying numbers of agents and diverse environmental constraints.
\end{itemize}

Our framework enables safety through geometric collision penalties, including self-collision modeling, while ensuring smooth, dynamically consistent trajectories through quadratic control regularization. The game-theoretic framework naturally models interaction between manipulators, enabling coordinated behavior to emerge from the optimization process without relying on fixed priority rules. By extending ILQ games to high-dimensional articulated systems, ILQ-Arm provides a principled and scalable solution for safe multi-agent manipulation in shared workspaces.

% \vspace{10pt}
\section{RELATED WORK}
This section reviews prior work on MAMP, which spans centralized and decentralized approaches, learning-based strategies, trajectory optimization techniques, and game-theoretic formulations.
\subsection{Centralized and Decentralized Classical Approaches}
Classical approaches to multi-agent motion planning can be broadly categorized into centralized and decentralized paradigms, each offering different trade-offs between coordination and scalability. The centralized motion planning methods jointly optimize the composite configuration space of all agents, allowing global coordinated behavior \cite{karaman2011sampling, schulman2014motion}. While it is effective for a smaller scale, its complexity grows rapidly with the number of agents and system dimensionality, limiting scalability in high DoF manipulation settings. On the other hand, to improve scalability, decentralized approaches allow each agent to plan independently. Geometric methods such as velocity obstacles and reciprocal velocity obstacles compute collision-free motions in velocity space \cite{fiorini1998motion, van2008reciprocal}, while Model Predictive Control (MPC) \cite{kouvaritakis2016model} has been extended to multi-robot coordination \cite{alonso2017multi}. Although computationally efficient, these methods often rely on simplified geometric assumptions and may struggle in tightly coupled scenarios involving articulated manipulators with complex collision geometries.

\subsection{Data-Driven Methods}
Learning-based approaches have recently shown promising performance in high-dimensional multi-robot systems. Neural policies trained via imitation learning or reinforcement learning enable fast deployment after offline training \cite{levine2016end}. Safety-aware extensions incorporate shielding mechanisms, control barrier functions (CBFs), or Hamilton-Jacobi reachability (HJR) constraints to enhance robustness \cite{ames2016control, fisac2018general}. Neural approximations of HJR value functions further improve scalability \cite{bansal2017hamilton, chen2025nehmo}. Despite these advances, learning-based methods often depend on extensive training data, may lack interpretability, and can exhibit limited generalization outside the training distribution.

\subsection{Trajectory Optimization}
Trajectory optimization \cite{betts1998survey} provides a complementary framework for handling nonlinear dynamics and rich safety constraints by directly optimizing trajectories over a finite horizon. Methods such as CHOMP \cite{ratliff2009chomp}, TrajOpt \cite{schulman2014motion}, and ILQR \cite{tassa2012synthesis} optimize smoothness, dynamic feasibility, and collision-aware objectives within a unified formulation. These approaches are well-suited for high-dimensional systems and can incorporate control regularization and goal costs in a principled manner. However, most optimization-based MAMP methods treat other agents as dynamic obstacles within a single aggregated objective that can implicitly favor one agent's cost over another's, rather than explicitly modeling their strategic interactions.

\subsection{Game-theoretic Formulations}
Game-theoretic formulations provide a principled way to capture coupled decision-making among multiple agents. Differential games model each agent as minimizing its own objective while considering the strategies of other agents, whose actions affect the shared dynamics \cite{bacsar1998dynamic}. LQ games allow the computation of feedback Nash equilibrium via Riccati-type recursions \cite{engwerda2005lq}, and ILQ methods extend this approach to nonlinear systems by iteratively linearizing the dynamics and approximating the cost \cite{fridovich2020efficient, tassa2012synthesis}. These techniques have been successfully applied in domains such as autonomous driving and competitive robotics \cite{fisac2019hierarchical, williams2017information}, where strategic iteration is central. Nevertheless, the application of ILQ differential games to high-DoF robotic manipulators remains largely unexplored. Prior work primarily focuses on low-dimensional vehicles or simplified collision models, leaving open the challenge of integrating complex link geometries and self-collision constraints within a game-theoretic framework.

\subsection{Positioning of Our Work}
In this work, we bridge trajectory optimization and differential game theory by formulating multi-manipulator motion planning as a differential game. Instead of imposing fixed priorities or treating agents as passive obstacles, we compute time-varying feedback Nash strategies using an \textbf{iterative LQ game} procedure. By explicitly incorporating quadratic control regularization, goal costs, and self-collision penalties, our approach extends ILQ game methods to high-dimensional articulated systems and demonstrates their effectiveness for coordinated multi-agent manipulation.

% \vspace{10pt}
\section{METHOD}
This section presents ILQ-Arm, our approach for solving multi-agent motion planning tasks for robot arm manipulators using the ILQ differential game framework for centralized motion planning. We begin by introducing the problem definition, followed by the necessary notations and the details of our method.

\subsection{Problem Definition}
In an environment with N agents at time $t$, the state is denoted $x(t)$, while the control signal applied to the state for all agents ranging from $1$ to $N$ is symbolized as $u_{1:N}(t)$. After applying control to the state over a time interval $\Delta t$, the system evolves according to the following discrete-time system dynamics
\begin{equation}
    x(t+1) = x(t) + f(x(t), u_{1:N}(t))\cdot\Delta t
    \label{eq:system}
\end{equation}
where $f(\cdot)$ is the system dynamics, which contains the dynamics of all agents. In our problem, each agent is modeled as an $m$-DOF robot arm manipulator with configuration $q$ in the C-space $\mathcal{Q}$, where $q \in \mathcal{Q} \subseteq \mathbb{R}^m$. The joint velocities and accelerations for each arm are denoted by $\dot{q} \in \mathbb{R}^m$ and $\ddot{q} \in \mathbb{R}^m$, respectively. The control signal $u$ corresponds to the acceleration of each joint among all agents. Consequently, the global state $x \in \mathbb{R}^{2mN}$ and joint control for agent $i$, denoted by $u_i \in \mathbb{R}^{m}$, are defined as:
\begin{align}
    x(t) &=\{q_{1:m}, \dot{q}_{1:m}, \dots, q_{N:m}, \dot{q}_{N:m}\} \\
    u_i(t)&= \{\ddot{q}_{1}, \dots, \ddot{q}_{m}\}
    \label{eq:playermodel}
\end{align}

Each agent $i$ aims to minimize a cost functional $J_i$, which consists of a running cost $\ell_i$ and a termination cost $\phi_i$. The running cost $\ell_i$ depends on all states $x$ and the control signal $u$ over the time horizon until $T-1$. In contrast, the termination cost $\phi_i$ is evaluated only with the final state $x(T)$ at time $T$. Thus, the cost function can be expressed as
\begin{equation}
    J_i(x) = \sum^{T-1}_{t=0} \ell_i(x(t), u_i(t))\cdot \Delta t + \phi_i(x(T)), \forall i \in \{1, \dots, N\}
    \label{eq:totalcost}
\end{equation}

\subsection{Background}

\begin{algorithm}[hbt!]
\caption{ILQ Arm solver}
\label{alg:ILQ}
\begin{algorithmic}[1]

% \Require initial state $x(0)$, control strategies $\{\gamma^0_i\}_{i \in [N]}$, time horizon $T$, running costs $\{g_i\}_{i \in [N]}$
% \Ensure converged control strategies $\{\gamma^*_i\}_{i \in [N]}$

\For{$k = 1, 2, \dots$}
    \State $\xi^k \gets \{ x(0), \dots, x(T) \}$
where $x(t+1) = x(t) + f(x(t), u_{1:N}(t))\cdot \Delta t$
    \State $\{A(t), B_i(t)\} \gets$ system linearization $\xi^k$
    \State $\{l_i(t), Q_i(t), r_{ij}(t), R_{ij}(t)\} \gets$ cost quadratic $\xi^k$
    \State $\{\tilde{\gamma}^k_i\} \gets$ solve LQ game $A(t), B_i(t), l_i(t),$ \\ $Q_i(t), r_{ij}(t), R_{ij}(t)$
    \State $\{\gamma^k_i\} \gets$ strategy update $\{\gamma^{k-1}_i, \tilde{\gamma}^k_i\}$
    \If{converged}
        \State \Return $\{\gamma^k_i\}$
    \EndIf
\EndFor

\end{algorithmic}
\end{algorithm}

Before delving into the implementation details, we first outline the overall design framework. Our goal is to compute a global Nash equilibrium in the form of time-varying state-feedback control strategies $\gamma_i^* \in \Gamma_i$, as defined in Eq.~\ref{eq:playermodel} and \ref{eq:totalcost}. For each agent $i$, the strategy space $\Gamma_i$ consists of feedback policies that map time and state to control inputs
\[
\gamma_i : [0,T] \times \mathbb{R}^{2mN} \rightarrow \mathbb{R}^{m_i}
\]

A set of strategies $\{\gamma_i^*\}_{i=1}^N$ constitutes a global Nash equilibrium if no agent can reduce its individual cost by unilaterally deviating. Formally, this is expressed as
\begin{equation}
\begin{aligned}
J_i^* 
&= J_i(\gamma_1^*, \dots, \gamma_i^*, \dots, \gamma_N^*) \\
&\leq 
J_i(\gamma_1^*, \dots, \gamma_{i-1}^*, \gamma_i, \gamma_{i+1}^*, \dots, \gamma_N^*) 
\quad \forall i, \; \forall \gamma_i 
\end{aligned}
\end{equation}

Computing a true global Nash equilibrium for nonlinear dynamics is generally intractable. Therefore, in practice, we seek solutions that satisfy the Nash conditions under local approximations of the dynamics and cost, leading to an ILQ game formulation.

We assume that each agent observes the full system state at every time step but does not have access to the control policies of other agents. 

The resulting procedure for LQ game, summarized in \cref{alg:ILQ}, begins with an initial state $x(0)$ and a set of initial feedback strategies $\{\gamma_i(0)\}_{i=1}^N$. These are run through non-linear dynamics $f(\cdot)$ to obtain the current trajectory $\xi^k$ at iteration $k$ (Line 2-3). To update the trajectory, we compute a Jacobian linearization of system dynamics $f$. We define the deviation from this trajectory $\delta x(t) = x(t)-\hat x(t)$ and $\delta u(t) = u_i(t)-\hat u_i(t)$ for each agent $i$. Using this, we can approximate the linearized system (Line 4) by
\begin{equation}
    \dot{\delta}x(t) \approx A(t)\delta x(t) + \sum_{i \in \{0,\dots,N\}} B_i(t)\delta u_i (t)
\end{equation}
where $A(t)$ and $B(t)$ are the Jacobian of state and control dynamics.

Next, we compute a quadratic approximation of the cost function $J_i$ as follows
\begin{equation}
\begin{aligned}
    \ell_i(&x(t), u_i(t)) \approx \\
    & \ell_i(\hat{x}(t), \hat{u_i}(t)) + \frac{1}{2} \delta x(t)^T(Q_i(t)\delta x(t) + 2l_i(t)) \\
    &+ \frac{1}{2} \sum_{j\in\{0,\dots,N\}} \delta u_j(t)^T (R_{ij}(t) \delta u_j(t) + 2 r_{ij}(t))
\end{aligned}
\end{equation}

\begin{equation}
\begin{aligned}
    \phi_i(&x(T)) \approx \phi_i(\hat{x}(T)) + \frac{1}{2} \delta x(T)^T \left( Q_i(T) \delta x(T) + 2l_i(T) \right)
\end{aligned}
\end{equation}
where $Q_i(t)$ and $R_i(t)$ are the Hessians $\nabla_{\hat{x}\hat{x}}^2 \ell_i$, $\nabla_{\hat{u}\hat{u}}^2 \ell_i$ and $l_i(t)$ and $r_{ij}(t)$ are the gradients $\nabla_{\hat{x}} \ell_i$, $\nabla_{\hat{u_j}} \ell_i$ (Line 5). For the goal cost, a similar approach is used, but it is computed only at the final time step $T$. Since there is no control input $u$ at the final step, the goal cost only depends on the final state deviation $\delta x(T)$.

We then solve the finite horizon LQ game, which returns a set of feedback strategies $\tilde{\Gamma}^k$ that constitute a Nash equilibrium for the LQ game. These optimal strategies are affine maps of the state deviation $\delta (t)$
\begin{equation}
    \tilde{\gamma}_i^k(t, x(t)) = \hat{u}_i(t) - P_i^k(t) \delta x(t) - \alpha_i^k(t)
\end{equation}
where $P_i^k(t)$ is the feedback gain matrix and $\alpha_i^k(t)$ is an affine term. To compute $P_i^k(t)$ and $\alpha_i^k(t)$, the algorithm starts at the final time $T$ and iterates backward to $t=0$. At each time step, it solves the equilibrium of the local LQ sub-game. Since it is a general-sum game, the agents' strategies are coupled and solved with coupled Riccati equations.

However, updating strategies often leads to divergence. The LQ solution assumes the linear dynamics and quadratic costs are accurate everywhere, but for a manipulator, these approximations are only valid locally (near the current trajectory $\xi^k$). To address this, the algorithm introduces a step size $\eta \in (0, 1]$. The actual strategy used in the update is given by (Line 8)
\begin{equation}
    \gamma_i^k(t, x(t)) = \hat{u}_i(t) - P_i^k(t) \delta x(t) - \eta \alpha_i^k(t)
\end{equation}
With the new strategies $\gamma_i^k$, we compute the next trajectory $\xi^{k+1}$ and repeat until convergence.

\subsection{Our method}
In this subsection, we present the details of our design, including the collision avoidance, the running cost formulation, and the termination cost design, as specified in Eq.~\ref{eq:totalcost}. We will then demonstrate how these components are integrated and reveal the final pipeline for solving multi-agent motion planning using our method.

%\textit{Maybe split into running cost vs terminal cost?}{\color{red}Yes, lets do that.}

% cost function design?
% 1. state: collision constraints (self, agent, static env, moving env)
% 2. state: goal position
% 3. control: effort function
\subsubsection{Collision Penalty}
Our method incorporates penalties for self-collision, inter-agent collisions, and environmental obstacles to ensure valid motion planning. Unlike point-mass models used in previous work \cite{fridovich2020efficient}, robot arm manipulators require sophisticated state-dependent cost terms to enable both the safety and validity of the generated path. To prevent self-collision in an $m$-DOF manipulator with $m+1$ links, we enforce a minimal distance between all non-adjacent link pairs $(l_i, l_j)$, ensuring it is greater than a safety margin $d_s$. The self-collision penalty $\sigma_s$ is formulated as follows, where $\lambda$ denotes the associated weighting coefficient 
\begin{equation}
    \sigma_s(x, d_{s}) = \lambda_s \min_{\substack{(i, j) \in \{m+1\}\times \{m+1\} \\ i - j > 1}}
    \left(  d_s - d_{min}(l_i, l_j)\right)
    \label{eq:self_collision}
\end{equation}
\par
For inter-agent collisions, we model the area occupied by all links of agent $i$ as a capsule volume $c_i$. The inter-agent collision penalty ensures that the minimal distance between any two agents is larger than the safety margin, which can be expressed as
\begin{equation}
    \sigma_t(x, d_s) = \lambda_t \min_{\substack{(i, j) \in \{N\}\times \{N\} \\ i \neq j}}
    \left(d_s - d_{min}(c_i, c_j)\right)
    \label{eq:inter_collision}
\end{equation}

To account for environmental obstacles, we utilize a Signed Distance Function (SDF) \cite{sethian1999level} to evaluate collision penalty
\begin{equation}
    \sigma_i(x, d_s) = \lambda_i \min_{\substack{i \in \{N\}}}
    \left( d_s - d_{sdf}(\{c_e\}, c_i) \right)
    \label{eq:env_collision}
\end{equation}
where $d_s(t, c_i):= \text{dist}_{\text{sdf}}(t, c_i)$ represents the distance between the manipulator and the workspace obstacles at time $t$. This formulation allows the ILQ solver to generate feasible multi-agent motion plans even in dynamic environments.

\subsubsection{Control Cost}
The quadratic control cost $\tilde{\iota}$ promotes smooth and physically feasible trajectories by penalizing excessive actuation. By minimizing the squared $L_2$-norm of the control input, the optimizer is encouraged to distribute effort over the entire horizon instead of producing impulsive, high-amplitude commands
\begin{equation}
    \tilde{\iota}_i(u) = \lambda_c \sum_0^{T-1} \|u_{0:m}\|_2^2
\end{equation}
Since the cost is strictly convex with respect to $u$, large control magnitudes are penalized disproportionately, suppressing abrupt variations in the input sequence. This indirectly yields smoother state evolution through the system dynamics. For robotic manipulators, this translates to fluid joint motions, reduced excitation of unmodeled dynamics, and improved mechanical stability.

From an optimization standpoint, the quadratic control term acts as a regularizer. It contributes positive definiteness to the control Hessian, improving numerical conditioning and enabling reliable backward-pass updates in second-order methods. This stabilizes convergence and prevents ill-posed updates associated with poorly conditioned curvature.

\begin{table*}[!ht]
    \vspace{0.1in}

  \fontsize{8}{10}\selectfont
  \begin{center}
    \begin{tabular}{c |c c c | c c c |c c c |c c c }
      \toprule
      \multirow{2}{*}{\textbf{Test Scenarios}}& \multicolumn{3}{c|}{\textbf{ILQ-Arm (Ours)}} & 
      \multicolumn{3}{c|}{\textbf{Centralized CHOMP}} &
      \multicolumn{3}{c|}{\textbf{Best Response CHOMP}} &
      \multicolumn{3}{c}{\textbf{Centralized RRT*}} \\
       & SR $\uparrow$ & PT $\downarrow$ & PL $\downarrow$ &
       SR $\uparrow$ & PT $\downarrow$ & PL $\downarrow$ &
       SR $\uparrow$ & PT $\downarrow$ & PL $\downarrow$ &
       SR $\uparrow$ & PT $\downarrow$ & PL $\downarrow$ \\
       \midrule
        2 agents & \textbf{96.8} & \textbf{0.121} & \textbf{10.109}
                & 89.2 & 5.777 & 34.331
                & 44.0 & 94.725 & 34.362
                & 62.8 & 15.110 & 23.455\\
        3 agents & \textbf{87.6} & \textbf{0.866} & \textbf{14.060}
                & 80.4 & 17.659 & 31.899
                & 26.0 & 130.511 & 28.616
                & 56.8 & 15.105 & 16.794\\
        4 agents & \textbf{86.0} & \textbf{1.954} & \textbf{15.767}
                & 62.0 & 44.726 & 33.039
                & 42.4 & 115.353 & 32.445
                & 40.8 & 15.101 & 19.921\\
        2 agents w obs & 95.2 & \textbf{0.585} & \textbf{7.893}
                & \textbf{98.8} & 1.698 & 12.78
                & 73.6 & 45.425 & 10.541
                & 87.6 & 15.103 & 8.665\\
       
      \bottomrule
    \end{tabular}
    \vspace{0.1in}

    \begin{tabular}{c |c c c | c c c |c c c |c c c }
      \toprule
      \multirow{2}{*}{\textbf{NeHMO Scenarios}}& \multicolumn{3}{c|}{\textbf{ILQ-Arm (Ours)}} & 
      \multicolumn{3}{c|}{\textbf{Centralized CHOMP}} &
      \multicolumn{3}{c|}{\textbf{NeHMO}} &
      \multicolumn{3}{c}{\textbf{Centralized RRT*}} \\
       & SR $\uparrow$ & PT $\downarrow$ & PL $\downarrow$ &
       SR $\uparrow$ & PT $\downarrow$ & PL $\downarrow$ &
       SR $\uparrow$ & PT $\downarrow$ & PL $\downarrow$ &
       SR $\uparrow$ & PT $\downarrow$ & PL $\downarrow$ \\
       \midrule
        2 agents & \textbf{90.8} & \textbf{0.217} & 12.387
                & 36.0 & 2.896 & 9.146
                & 77.0 & 5.304 & 30.442
                & 55.2 & 15.107 & 16.799\\
        3 agents & \textbf{86.8} & \textbf{0.455} & 17.042
                & 20.0 & 6.714 & 13.156
                & 80.0 & 4.025 & 36.758
                & 42.4 & 15.134 & 19.499\\
        % 4 agents & 86.6 & 0.704 & 15.812
        %         & 62 & 44.726 & 33.039
        %         & 42.4 & 115.353 & 32.445
        %         & - & - & -\\
       
      \bottomrule
    \end{tabular}
    \caption{This table compares ILQ-Arm with the baseline methods. ILQ-Arm achieves the best overall performance in success rate, planning time, and path length. Centralized CHOMP performs reasonably well but is computationally expensive and scales poorly as the number of agents increases. NeHMO achieves moderate success rates but produces less efficient solutions with longer paths. In contrast, ILQ-Arm consistently scales to complex scenarios while maintaining high success rates and the lowest planning times, demonstrating its effectiveness for multi-agent rearrangement.}
    \label{tab:baseline}
  \end{center}
  \vspace{-0.2in}
\end{table*}

\subsubsection{Running Cost}
The previously defined collision penalties and control cost are combined into the running cost $\ell_i$ from Eq.~\eqref{eq:totalcost} as follows
\begin{equation}
\begin{aligned}
    \ell_i(x, u_i) &= \sigma_s(x, d_s) + \sigma_t(x, d_s) \\
    &+ \sigma_i(x, d_s) + \tilde{\iota}_i(u)
\end{aligned}
\end{equation}

\subsubsection{Termination Cost}
Our termination cost $\phi(x)$ is defined as a quadratic penalty on the final state deviation in order to ensure goal reachability within a finite-time horizon. This formulation encourages the system to minimize the error between the terminal state and the desired configuration $x^g$
\begin{equation}
    \phi(x) = \frac{1}{2} \beta \|x^g_i - x_i(T)\|_2^2
\end{equation}
where $x^g \in \mathcal{Q} \subseteq \mathbb{R}^m$ denotes the target joint configurations for agent $i$, $x_i(T)$ is the joint angles at the final time $T$, and $\beta$ is a scalar weight that determines the priority of goal attainment relative to other cost components.

While a running cost penalizes goal deviation throughout the entire trajectory, it often biases the solver toward a direct path, which can lead to local minima or infeasible solutions when complex obstacles are present. In contrast, defining $\phi(x)$ as a terminal-only penalty provides flexibility to the planner. By isolating the goal objective from the intermediate states, the optimizer can focus on collision avoidance and dynamic feasibility during the transient phase. This allows the robot to make detours or execute wide maneuvers to bypass obstacles without incurring a high cost penalty, provided it converges to the target configuration $x^g$ by the final time $T$.

The coupling between the termination cost and the running cost naturally induces braking behavior near the goal state $x^g$. As the system approaches the target, aggressive control inputs become costly, encouraging gradual deceleration and reducing overshoot. This results in a trajectory that balances goal achievement with motion quality while maintaining dynamic consistency.

\begin{figure}[t]
\vspace{0.1in}
\centering

\includegraphics[trim={0.6cm 0.9cm 0.6cm  0.8cm},clip, width=0.9\linewidth]{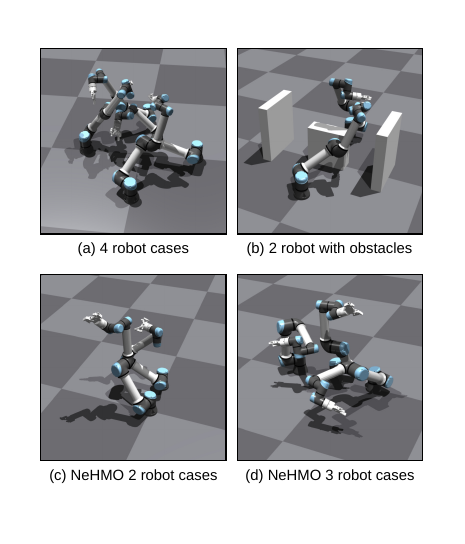}

\caption{This figure illustrates representative test scenarios, including the 4-agent setup, the 2-agent case with obstacles, and the NeHMO benchmarks with 2 and 3 agents.}
\label{fig:simulation}
\vspace{-0.2in}

\end{figure}

% \begin{figure}[t]
% \vspace{0.1in}
% \centering
% \begin{subfigure}{0.23\textwidth}
% \centering
% \includegraphics[trim={0cm 0cm 0cm  0cm},clip, width=0.9\linewidth]{images/rob4.png}
% \caption{4 robot case}
% \end{subfigure}
% \begin{subfigure}{0.23\textwidth}
% \centering
% \includegraphics[trim={0cm 0cm 0cm 0cm},clip,width=0.9\linewidth]{images/obs.png}
% \caption{2 robot with obstacles}
% \end{subfigure}
% \begin{subfigure}{0.23\textwidth}
% \centering
% \includegraphics[trim={0cm 0cm 0cm 0cm},clip,width=0.9\linewidth]{images/nehmo_3.png}
% \caption{NeHMO 3 robot case}
% \end{subfigure}
% \begin{subfigure}{0.23\textwidth}
% \centering
% \includegraphics[trim={0cm 0cm 0cm 0cm},clip,width=0.9\linewidth]{images/nehmo_2.png}
% \caption{NeHMO 2 robot case}
% \end{subfigure}

% \caption{This figure illustrates representative test scenarios, including the 4-agent setup, the 2-agent case with obstacles, and the NeHMO benchmarks with 2 and 3 agents.}
% \label{fig:simulation}
% \vspace{-0.2in}

% \end{figure}

% \vspace{10pt}
\section{EXPERIMENTS}
In this section, we present the simulation setup, the
baseline comparison, the ablation studies, and the real-world
experiments.

\subsection{Simulation Setup}
We construct six types of complex simulation scenarios to evaluate the performance of our method against the baselines in multi-agent systems operating within cluttered environments. For each scenario, we randomly generate 250 test instances for comprehensive evaluation, with $T = 5$s, $\Delta t = 0.1$s, and convergence threshold of $1 \times 10^{-1}$. All test cases are restricted to instances in which collisions would occur if each agent moved directly toward its target configuration.

In Scenarios 1-4, manipulators were positioned uniformly along a circle with a radius of 0.6 m, with the number of agents varying from two to four. Each environment includes a ground plane as a static obstacle. Scenario 4 also incorporated multiple static obstacles to assess environmental collision-avoidance performance in confined, narrow scenes.

Scenarios 5 and 6 are designed to enable direct comparison with the pretrained model proposed by NeHMO \cite{chen2025nehmo}, ensuring consistency in evaluation conditions. In total, 1,500 test cases were generated for multi-agent path planning. Examples of testing instances in diverse environments are shown in Fig. \ref{fig:simulation}.

To quantitatively assess performance, we adopted the following evaluation metrics.
\begin{itemize}
\item \textbf{Success Rate (SR)}: Success rate is defined as the percentage of valid plans in which all agents reach their respective goal configurations without collisions. A trial is considered a failure if it fails to complete within 120 seconds or fails to converge within 200 iterations.
\item \textbf{Planning Time (PT)}: Planning time measures the total computational time required for the planner to generate feasible trajectories.
\item \textbf{Path Length (PL)}: Path length represents the cumulative travel distance in radians of all agents, reflecting the quality and efficiency of the generated trajectories.
\end{itemize}

\subsection{Simulation Results}
\subsubsection{Baseline Comparison}
To evaluate the effectiveness of the proposed approach, we compare it against the following three baseline methods.
\begin{itemize}
    \item \textbf{Centralized, CHOMP}: The system is centralized by concatenating the state vectors of all agents into a single composite state space, which then uses CHOMP \cite{ratliff2009chomp} to solve the planning problem.
    \item \textbf{Best Response, CHOMP}: Each agent generates its trajectory sequentially while treating the previously planned trajectories of other agents as obstacles. This process is repeated until convergence, defined as no further updates to any agent’s trajectory.
    \item \textbf{Centralized, RRT*}: The centralized formulation is identical to Centralized CHOMP, but RRT* is used as the planner instead of CHOMP. Since this method continually optimizes the plan as time allows, we set the time limit to a reasonable 15 seconds.
    \item \textbf{NeHMO}: A decentralized learning-based motion planner that leverages Hamilton–Jacobi reachability to ensure inter-agent collision avoidance \cite{chen2025nehmo}. However, NeHMO cannot handle obstacles. Thus, to ensure a fair comparison, we evaluate our method against it using the configurations reported in the original paper, shown in Table \ref{tab:baseline} as the NeHMO scenarios.
\end{itemize}

Table \ref{tab:baseline} reports results across the different test scenarios with all metrics averaged over successful instances except for success rate.

In our test scenarios, Centralized CHOMP performs reasonably but is outperformed by ILQ-Arm in both efficiency and path quality. In the test cases with static obstacles, Centralized CHOMP achieves slightly higher success rates than ILQ-Arm. This can be attributed to its centralized formulation, which plans over the full joint state space of all agents simultaneously, inherently avoiding collisions in constrained or obstacle-rich configurations. However, this advantage comes at a significant cost: planning times increase dramatically, and the resulting paths are often much longer and less efficient. Thus, while it may succeed in avoiding collisions in highly constrained environments, it does so inefficiently and lacks scalability as the number of agents increases.

Decentralized CHOMP consistently struggles across all scenarios. Its sequential, best-response approach fails to maintain high success rates in both random and complex environments, particularly as agent numbers grow. Planning times are extremely high, and the generated trajectories are generally inefficient as we can see from path length data. Compared to ILQ-Arm, it demonstrates both poor reliability and poor computational efficiency.

NeHMO is a decentralized, learning-based planner that uses Hamilton–Jacobi reachability to avoid inter-agent collisions. It achieves relatively high success rates in the given NeHMO scenarios due to its effective inter-robot collision avoidance. However, its performance is limited in several key ways. Planning times are significantly higher than those of ILQ-Arm, and the generated trajectories are substantially longer due to its reachability modeling, reflecting inefficiency in path quality. Moreover, NeHMO cannot account for static obstacles, which restricts its applicability in cluttered or confined environments. As the number of agents increases, success rates decline further, and path inefficiency becomes more pronounced, making it unsuitable for larger or more complex tasks.

\begin{figure*}[t]
\vspace{0.1in}
\centering
    \includegraphics[trim={0.6cm 0.0cm 0.6cm 0.4cm},clip,width=17.5cm]{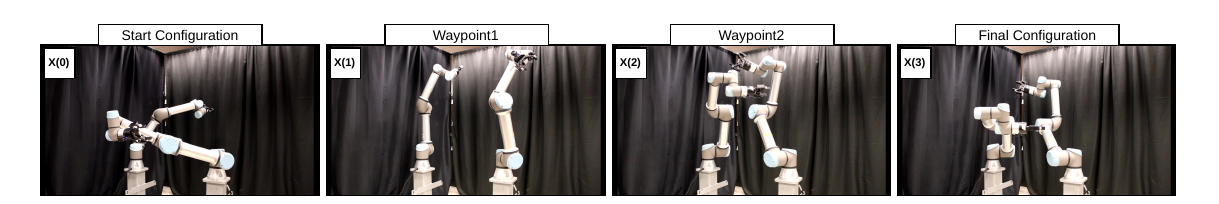}
    %\centering
    \caption{The figure shows a multi-agent motion planning. Each key-frame illustrates the system’s motion, with the corresponding time step indicated in the upper-right corner. At way-points 1 and 2, the robots demonstrate effective inter-robot collision avoidance while simultaneously optimizing their trajectories toward the final configuration. Our ILQ-Arm finds collision free solution in 0.05 seconds, while the execution lasted 45 seconds.
    }%
    \label{fig:real_exp2}%
    \vspace{-0.1in}
\end{figure*}

\begin{table*}[!ht]
  \fontsize{8}{10}\selectfont
  \begin{center}
    \begin{tabular}{c |c c c c | c c c |c c c |c c c }
      \toprule
      \multirow{2}{*}{\textbf{Test Scenarios}}&
      \multicolumn{4}{c|}{\textbf{ILQ-Arm (Ours)}} & 
      \multicolumn{3}{c|}{\textbf{ILQ-Arm w/o self-collision}} &
      \multicolumn{3}{c|}{\textbf{ILQ-Arm w Pseudo-Huber}} &
      \multicolumn{3}{c}{\textbf{ILQ-Arm w running goal}} \\
       & SR $\uparrow$ & \#CL $\downarrow$ & PT $\downarrow$ & PL $\downarrow$ &
       \#CL $\downarrow$ & PT $\downarrow$ & PL $\downarrow$ &
       SR $\uparrow$ & PT $\downarrow$ & PL $\downarrow$ &
       SR $\uparrow$ & PT $\downarrow$ & PL $\downarrow$ \\
       \midrule
        2 agents & \textbf{86.4} & \textbf{0} & 0.659 & 7.668
                & 31 & \textbf{0.0950} & \textbf{7.649}
                & \textbf{86.4} & 0.656 & 7.668
                & 74.4 & 1.689 & 9.154 \\
        3 agents & \textbf{78.4} & \textbf{0} & 1.734 & 14.206
                & 17 & \textbf{1.123} & 14.177
                & 74.0 & 1.713 & \textbf{14.177}
                & 25.2 & 7.408 & 15.160 \\
        4 agents & \textbf{84.0} & \textbf{0} & 2.287 & \textbf{14.906}
                & 16 & \textbf{1.466} & 14.907
                & 82.8 & 2.781 & 14.909
                & 37.6 & 8.414 & 17.590 \\
        2 agents w obs & \textbf{86.4} & \textbf{0} & 1.343 & \textbf{7.910}
                & 19 & \textbf{0.545} & 7.943
                & \textbf{86.4} & 1.304 & \textbf{7.910}
                & 67.6 & 1.480 & 9.277 \\
       
      \bottomrule
    \end{tabular}
    \vspace{0.1in}
    \caption{
       The ablation study highlights the importance of the control cost, terminal goal cost, and self-collision penalty in multi-agent motion planning. Removing the self-collision term reduces computation time but eliminates collision-free guarantees, making the planner unsafe for real-world use. Overall, the full ILQ-Arm formulation achieves the best balance of success rate, collision avoidance, planning time, and path length. These results demonstrate that all cost components are necessary for reliable, efficient, and dynamically consistent motion planning.
    }
    \label{tab:ablation}
  \end{center}
  \vspace{-0.2in}
\end{table*}

In conclusion, ILQ-Arm achieves the best overall balance of success rate, planning time, and path quality. Even in complex environments, it produces far shorter and more efficient paths in a fraction of the time. This demonstrates that ILQ-Arm not only scales effectively with agent number and environment complexity but also maintains high efficiency, making it practical and robust for multi-agent motion planning tasks.

\subsubsection{Ablation Studies}
We conduct ablation studies to validate the effectiveness of our design choices and to analyze performance sensitivity on manipulator systems. The following variants are evaluated.
\begin{itemize}
    \item \textbf{ILQ-Arm without self-collision penalty}: In this variant, the self-collision penalty is removed to evaluate its significance. Previous models often ignored self-collision due to their simplicity or 2D assumptions, but for high-DoF manipulators, this component may critically impact safety and trajectory feasibility.
    \item \textbf{ILQ-Arm with Pseudo-Huber control cost}: We replaced quadratic control with a Pseudo-Huber loss. Since the Pseudo-Huber function transitions from quadratic behavior near the origin to approximately linear growth for larger inputs, this experiment examines how altering the control penalty structure affects convergence and trajectory behavior.
    \item \textbf{ILQ-Arm with running goal cost}: The goal cost is formulated as a running cost $\ell_i$, rather than a terminal cost $\phi_i$. This will continuously penalize deviations from motion toward the goal throughout the trajectory.
\end{itemize}

The quantitative results are summarized in Table \ref{tab:ablation}, where CL denotes the total number of self-collisions in the final trajectories. These test cases are more complicated than those used in Table \ref{tab:baseline}, as we want to comprehensively reveal the performance differences between the variations.

% We created more complex test cases than those used in the baseline comparison to clearly distinguish the performance difference between each models.

From Table \ref{tab:ablation}, we can see that removing the self-collision penalty significantly reduces planning time across all scenarios due to the simplified objective. However, it leads to frequent collisions. This confirms that, although computationally cheaper, the planner becomes unsafe and unreliable without explicit self-collision modeling.

In addition, replacing the quadratic control cost with a Pseudo-Huber loss produces comparable success rates and path lengths in simpler settings, such as the 2-agent case. However, as the number of agents increases, the quadratic formulation consistently achieves higher success rates and more stable convergence. This suggests that the stronger curvature of the quadratic penalty improves numerical conditioning and stabilizes the backward pass in the LQ game iterations.

Finally, formulating the goal cost as a running term substantially degrades performance. Success rates drop sharply in multi-agent scenarios, and both planning time and path length increase. This indicates that distributing goal pressure uniformly across the horizon interferes with coordinated multi-agent negotiation, whereas a goal formulation better preserves flexibility during intermediate motion.

Overall, the full ILQ-Arm formulation achieves zero collisions while maintaining high success rates, competitive planning times, and short path lengths across all test scenarios. These results demonstrate that each cost component plays a complementary role, and their combination yields the most balanced and robust performance.

\subsection{Real Robot Experiments}
We deploy our ILQ-Arm on a multi-agent system composed of two UR5e robots placed 0.8 m apart from each other. We conducted a total of four inter-robot collision avoidance tasks and one with the static obstacles to verify the real-world performance of our approach. One example can be found in Fig. \ref{fig:real_exp1}, where  two robots navigate through narrow environments avoiding static obstacles and inter-robot collisions. Another example is shown in Fig. \ref{fig:real_exp2}. The two robots avoid inter-robot collision while optimizing their own cost function. During the experiment, our method succeeds in solving all cases with an average planning time of 0.475 seconds and an average execution time of 64 seconds. In the generated plan, the robots perform synchronized actions for each time step by following the solved states, successfully avoiding all collisions, and eventually reaching the goal.

\section{CONCLUSIONS}
This paper presents a differential game framework for multi-agent motion planning among multiple high-DoF robotic manipulators. By formulating multi-arm coordination as an ILQ game, the proposed method computes time-varying state feedback Nash strategies that naturally capture the strategic coupling among manipulators. The framework integrates differentiable self-collision and inter-arm collision penalties directly into the second-order optimization pipeline, enabling safe and coordinated motion generation without imposing fixed priority rules. Through extensive simulation studies on challenging multi-arm scenarios, we demonstrate that the approach produces smooth, dynamically consistent, and collision-aware trajectories while maintaining computational efficiency. Compared with conventional centralized optimization and decoupled planning strategies, the game-theoretic structure better captures interactive behaviors and leads to more balanced coordination among agents.

\bibliographystyle{unsrt}
\bibliography{root}

\end{document}